\documentclass[twocolumn]{article}
\usepackage{amsmath,amssymb,amsfonts}
\usepackage{graphicx}
\usepackage{textcomp}
\usepackage{xcolor}

\usepackage{url}
\usepackage{csquotes}
\usepackage{array}
\usepackage{svg}
\usepackage{amsfonts}
\usepackage{amsmath}
\usepackage{breqn}
\usepackage{diagbox}
\usepackage{float}
\usepackage{caption}
\usepackage{subcaption}
\usepackage{rotating}
\usepackage[margin=1in]{geometry} 
\usepackage[ruled,vlined]{algorithm2e}
\usepackage{array}
\newcolumntype{P}[1]{>{\centering\arraybackslash}p{#1}}
\newcolumntype{M}[1]{>{\centering\arraybackslash}m{#1}}
\graphicspath{ {./files/} }
\usepackage[
backend=biber,
style=numeric,
sorting=none
]{biblatex}
\def\BibTeX{{\rm B\kern-.05em{\sc i\kern-.025em b}\kern-.08em
    T\kern-.1667em\lower.7ex\hbox{E}\kern-.125emX}}

\usepackage{enumitem} 
\usepackage{amssymb} 

\begin{document}

\title{Design Principles for Falsifiable, Replicable and Reproducible Empirical ML Research}

\author{
    Daniel Vranješ \\
    \textit{Helmut Schmidt University} \\
    \texttt{daniel.vranjes@hsu-hh.de}
    \and
    Oliver Niggemann \\
    \textit{Helmut Schmidt University} \\
    \texttt{oliver.niggemann@hsu-hh.de}
}
\date{} 

\maketitle

\begin{abstract}
Empirical research plays a fundamental role in the machine learning domain. At the heart of impactful empirical research lies the development of clear research hypotheses, which then shape the design of experiments. The execution of experiments must be carried out with precision to ensure reliable results, followed by statistical analysis to interpret these outcomes. This process is key to either supporting or refuting initial hypotheses. Despite its importance, there is a high variability in research practices across the machine learning community and no uniform understanding of quality criteria for empirical research. To address this gap, we propose a model for the empirical research process, accompanied by guidelines to uphold the validity of empirical research. By embracing these recommendations, greater consistency, enhanced reliability and increased impact can be achieved.
\end{abstract}


\section{Introduction} \label{introduction}

Deductive and abductive reasoning play crucial roles in both theoretical and empirical research. Deductive reasoning, often used in theoretical research, involves deriving specific predictions from general principles or axioms. This approach ensures internal consistency within a logical framework but does not require empirical testing. In contrast, abductive reasoning, which is foundational in empirical research, involves forming plausible hypotheses to explain observations. These hypotheses are then tested through experiments to gather empirical evidence. While deductive reasoning provides a solid foundation for developing theoretical models, abductive reasoning bridges the gap between theory and practice by enabling the formulation and empirical validation of testable hypotheses.

Theoretical research in machine learning (ML) involves the development and analysis of models and algorithms through mathematical formalisms and proofs, offering insights into their properties, performance guarantees, and limitations. This foundational work is essential for understanding the principles that govern ML systems, guiding the design of new algorithms, and providing a basis for interpreting their behavior. The analytical and formal nature of this approach allows for certain and specific evidence, but is not applicable for most hypotheses.

Empirical research, on the other hand, plays a crucial role in the ML domain through testing and validating of theoretical models and hypotheses in practical settings where theoretical proof is not possible \cite{Keuth2007}.

ML is inherently more empirical and experimental compared to symbolic AI due to its reliance on data-driven methods to develop and refine algorithms. Unlike symbolic AI, which relies on predefined rules and logic, ML systems learn patterns and make decisions based on large datasets. This necessitates extensive experimentation to evaluate the effectiveness, robustness, and generalizability of models across diverse and often unseen data. The iterative process of training, testing, and validating ML models requires empirical evidence to ensure their accuracy and reliability in real-world applications, making experimentation a cornerstone of the ML research methodology.

Despite the existence of abstract process models and guidelines for the operational aspects of ML (MLOps), the scientific ML community lacks standard process models for conducting research and even a common conception of what good research practices are in general and in detail. Empirical ML research faces several critical challenges that can impact its scientific rigor and the applicability of its findings. One significant issue is the problem of falsifiability, where some empirical studies may not clearly delineate conditions under which the proposed models could be proven wrong, leading to ambiguous interpretations of results. Replicability and reproducibility are also major concerns. The former refers to the ability of different researchers to achieve the same results using the same dataset and methodology, while the latter pertains to the capacity to achieve consistent results across different settings, datasets, and experimental configurations. The complexity and variability inherent in ML models, combined with the use of proprietary datasets or software and insufficient documentation, often hinder these aspects. Furthermore, generalizability is a persistent problem, as models trained and tested on specific datasets or in controlled environments may not perform well on unseen data or in real-world applications, questioning the external validity of the research. These challenges underscore the need for rigorous methodological standards, transparent reporting, and comprehensive evaluation metrics in empirical ML research to ensure its findings are robust, reliable, and broadly applicable.

Within this paper we focus on the design and execution of experiments as one encapsulated part of the ML pipeline from a scientific research perspective without any emphasis on the overall aspects of the life cycle of ML software or organizational aspects as e.g. promoted by process models such as CRISP-ML(Q) \cite{Studer2020} or the ISO/IEC 23053:2022.

In reviewing diverse sources, it becomes evident that certain terminologies are employed with varying definitions. \cite{Gunderson2018} highlight, for instance, that the concepts of replicability and reproducibility lack uniformity in definition, with a plethora of terms and interpretations in circulation. Given the critical need for a shared lexicon in scholarly discourse, we propose the following definitions to clarify our interpretation of these key terms within this publication.

\textbf{Scientific empiricism:} Scientific empiricism is a philosophical approach that underscores the importance of empirical evidence in the acquisition and validation of knowledge. Within this framework, verificationism and falsification play crucial roles as methodologies for evaluating scientific theories. Verificationism seeks to confirm the validity of hypotheses by demonstrating that empirical evidence is consistent with them, highlighting the role of positive evidence in supporting scientific claims. On the other hand, falsification, popularized by Karl Popper \cite{Keuth2007}, argues that scientific theories cannot be conclusively verified but can be robustly tested by seeking evidence that could potentially refute them. This approach emphasizes the critical importance of disprovability as a hallmark of scientific knowledge, suggesting that theories must be open to empirical testing and possible rejection. Together, verification and falsification embody the empirical spirit of scientific inquiry, advocating for a rigorous and evidence-based approach to understanding the natural world. Verification and falsification can even be attributed with a methodological symmetry \cite{Rapp1975}. Research hypotheses define the aim of the research and represent it's evaluation criteria. To enable falsifiability or verification they must be precise and answerable within the scope of the conducted research. Experiments must be conducted in a way to generate data which allows for a statistically significant analysis of hypotheses and empirical evidence shall be used to prove or disprove the hypotheses. Ensuring inter-subjectivity in ML research is crucial, as it allows different researchers to independently reproduce and validate findings, thereby enhancing the reliability and credibility of the results. This collective verification process strengthens the overall robustness and acceptance of empirical research within the scientific community. This step requires the replicability or reproducibility of the research.

\textbf{Replicability:} This principle dictates that identical results should be generated when an experiment is rerun using the same data, code and computational environment (hardware and software). Conducting the experiment multiple times on the same machine should yield consistent results. Replication can be used to detect fraud \cite{Drummond2009} or to check for stochastic behavior of the program leading to arbitrary results over multiple runs. Here the ML method of interest is strictly bound to its implementation resulting in the evaluation of the concrete ML program. Replicability is the lowest level of independent verification of research through third parties. The replicability within the initial research team resembles only a repeatability, which allows for the evaluation of the stochastic behavior but does not increase validity with regards to the approval by the scientific community.

\textbf{Reproducibility:}
Reproducibility has a broader understanding of the recreation of results than replicability. In \cite{Gunderson2018} reproducibility is defined as "the ability of an independent research team to produce the same results using the same AI method based on the documentation made by the original research team". Here the recreation can differ in terms of code or computational environment used for the experiments. Results should still support the hypotheses of the initial research. Using the same experimental methods should lead to similar results in multiple studies. Since this is not bound to a specific implementation, reproducibility does not evaluate a specific ML program but a ML method \cite{Gunderson2018}.

\textbf{(Cross-Domain) Generalizability:}
In cyber-physical systems (CPS), significant variations in the underlying physical properties across different systems challenge the predictability of algorithm transferability. This variability necessitates careful consideration of the specific application data used in algorithm evaluations. To foster robust and generalizable hypotheses, it is imperative to treat the data or use case as a variable and conduct evaluations across a diverse array of use cases. Such an approach ensures that empirical findings are not confined to a single application, enhancing the broader applicability of the results. Furthermore, in tasks like anomaly detection or classification, assessing the cross-domain generalizability of new algorithms is crucial. This aspect can be viewed as an extreme form of reproducibility, where the algorithms must maintain performance despite significant shifts in data characteristics and context.

\section{State of the Art} \label{sota}

Researchers from different application domains within the ML community have already (partially) addressed attributes of good research as well as current problems of and possible solutions for such research.

From a theoretical perspective, the cornerstone of research across various fields is the development of research hypotheses. Central to this endeavor are universally applicable concepts such as inductive and deductive reasoning, along with the principles of verification and falsification \cite{Rapp1975, Holyoak2012}. Although no singular methodology reigns supreme, each approach fundamentally aims to craft research hypotheses that are subject to rigorous testing and validation through coherent argumentation.

On the applied side, as elaborated in \cite{Drummond2009} and \cite{Drummond2018} the majority of the scientific community views the replicability of research as desirable, whilst the author himself does not view the sole replicability as good research and in contrast promotes the importance of reproducibility. ML research has significant potential for improvement regarding it's replicability as stated in \cite{Hutson2018}, due to a high amount of research being published without code or data.

We randomly select 100 research papers from the the 2023 International Joint Conference On Artificial Intelligence (IJCAI). We look at papers \#1 to \#75 of the main track and papers \#378 - \#399 of the ML track and evaluate those that contain experimental research for methods applied to the experiments. We find the mentions of the following methods in multiple research papers. No paper has used all of the methods.

\textbf{Benchmark against baselines:}
Comparing the proposed algorithm with state-of-the-art competitors is fundamental. This involves identifying current leading algorithms in the field and benchmarking the proposed algorithm's performance against these. The comparison should be fair and transparent, using the same datasets, pre-processing techniques, and evaluation metrics. Hyperparameters shall be optimized for baselines as well. This helps in assessing the true innovation or improvement offered by the proposed algorithm.

\textbf{Using multiple data sets:}
Benchmarking on multiple diverse datasets is crucial to demonstrate the robustness and generality of the algorithm. Each dataset should represent different aspects of the problem domain, varying in size, complexity, and inherent biases. This approach ensures that the algorithm's performance is not tailored to the idiosyncrasies of a single dataset but is broadly applicable across the domain. Where applicable, commonly used benchmarks shall be used for evaluation to allow for comparison with state of the art alternatives.

\textbf{Data shuffling:}
Data shuffling is a preprocessing technique used to enhance the generalization ability of models by randomizing the order of data points in the training dataset. This process helps in mitigating the risk of learning spurious patterns that may arise from the order-dependent biases inherent in the data collection process.

By ensuring that each mini-batch of data, used during the training phase, contains a diverse set of examples, shuffling helps in preventing the model from memorizing the sequence of training examples, thereby improving its ability to generalize from the training data to unseen data. In ML tasks involving time series data, data shuffling can be used when the series has been sliced into windows or segments, allowing models to learn from a variety of temporal patterns without compromising the sequential integrity within each segment.

\textbf{Cross-validations:}
Cross-validation is a statistical method to assess how the results of an algorithm will generalize to an independent data set. It is primarily used to estimate the skill of a model on unseen data. This technique involves partitioning a sample of data into complementary subsets, performing the learning on one subset (called the training set), and validating the analysis on the other subset (called the validation set or testing set). The process is repeated multiple times, with different partitions, to reduce variability. Common forms of cross-validation are k-fold cross-validation and leave-one-out cross-validation.

\textbf{Hyperparameter optimization:}
Hyperparameters control the learning process and can significantly affect performance. Hyperparameter optimization seeks the optimal set of hyperparameters that yields the best performance on a validation set. Techniques such as grid search, random search, or more sophisticated methods like Bayesian optimization can be used. This process ensures that the algorithm's potential is fully explored.

\textbf{Controlling random seeds:}
Manually setting the values of random seeds for random number generators in experiments and varying these values tests the robustness of the results against stochastic variations. It helps to identify if the observed performance is genuinely attributable to the algorithm's efficacy or merely a result of random chance. This practice enhances the credibility of the results.

\textbf{Replication runs:}
Stochastic variations can arise from initial model parameter settings, randomness in data shuffling and partitioning, variability in the optimization process such as stochastic gradient descent, and differences in the allocation of computational resources. Experimental replication involves conducting the same experiment multiple times with identical configurations to assess the consistency of the results amidst inherent stochastic variations. This method is crucial in distinguishing between true experimental effects and outcomes influenced by random chance.

\textbf{Ablation studies:}
Ablation studies systematically remove parts of the algorithm (e.g., features, layers in a neural network) to understand the contribution of each component to the overall performance. This method helps in identifying the critical elements of the algorithm and can guide further refinement and simplification.

\textbf{Sensitivity analysis:}
Sensitivity analysis involves varying the algorithm's parameters or the configuration of its modules to assess the impact on performance. This analysis can reveal dependencies and robustness of the algorithm against changes, providing insights into its behavior under different conditions.

\textbf{Result averaging and confidence intervals:}
Averaging results across multiple runs (due to rerunning, different seeds, or cross-validations) and computing mean and standard deviation for selected metrics provides a more stable and reliable estimate of the algorithm's performance. This method helps in summarizing the results and facilitates comparison with other algorithms. Alongside mean performance metrics, reporting confidence intervals provides a statistical range within which the true performance metric is likely to lie. This gives a clearer picture of the variability and reliability of the performance estimates. A narrow confidence interval around the average result suggests high reproducibility and stability of the algorithm's performance, whereas a wide interval signals greater variability and potential sensitivity to experimental conditions.

\textbf{Statistical testing:}
Statistical tests can be employed to rigorously evaluate whether the differences in performance metrics between algorithms are statistically significant. This adds a layer of quantitative rigor to the analysis, supporting stronger conclusions about the comparative effectiveness of the algorithms.

\section{Experimental Process}

\begin{figure*}[ht]
\centerline{\includegraphics[scale=0.8]{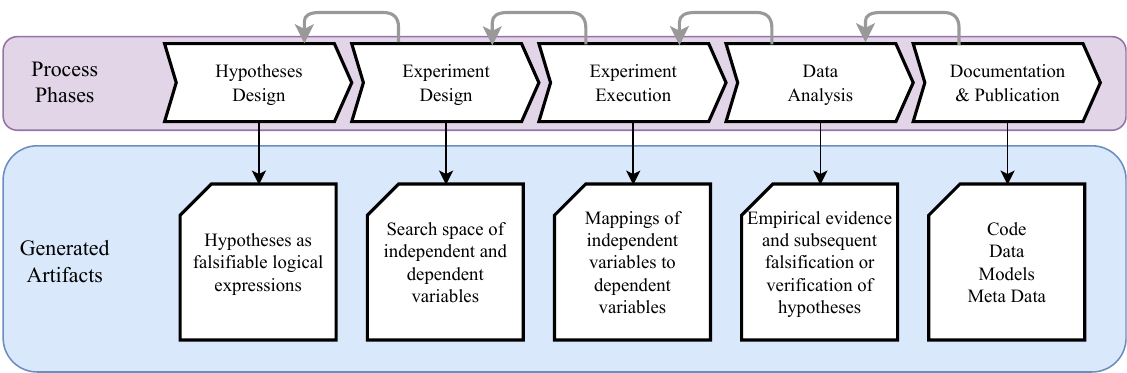}}
\caption{Research process model with process phases and generated artifacts}
\label{figure:process}
\end{figure*}

We define the experimental process (see fig. \ref{figure:process}) beginning with the formulation of research hypotheses. Based on the hypotheses, experiments are designed. The experimental design defines the necessary input and output of the experiment as well as the metrics to be applied to it to generate empirical evidence. The execution of the experiment generates data, which through an analysis generates evidence. This evidence is directly linked to the hypotheses and used for validation or falsification of them. Results are then documented and published. In the following we describe the phases in more detail and elaborate our proposed methods to increase empirical research quality. Whilst our research process model shows a linear forward oriented order of process steps, research is an iterative process. Cycles within this process are natural and a step back into a previous phase due to new insight does comply with our process model.

\subsection{Formulation of Hypotheses}
A hypothesis \cite{Bauer2022, Ishikawa2024} is a logical statement that postulates a specific relationship among the independent variables $X$, the control variables $C$, and the dependent variables $Y$ of an experiment. $X$ and $C$ represent the input variables of an experiment and $Y$ the output observations, see chapter \ref{expdes}. This relationship can be expressed in terms of equality, inequality, or other logical constructs that can be evaluated as true or false based on empirical data obtained from the experiment. The hypothesis can be categorized into two main types: null hypothesis $H_0$ and alternative hypothesis $H_a$.

The null hypothesis $H_0$ typically asserts that there is no effect or no difference, or it states a specific value or condition expected in the absence of a relationship among the variables. For example, $H_0$ could assert that the mean response is not affected by different levels of an independent variable. The use of null hypotheses provides a clear, objective baseline for comparison and ensures the rigor of scientific research by requiring falsifiable claims. It allows for the control of errors and the determination of statistical significance, thereby enhancing the reliability of the research findings.

The alternative hypothesis $H_a$ proposes a relationship or effect that contradicts the null hypothesis. It is usually what the research seeks to support through the data. For instance, $H_a$ might suggest that the mean response for one level of an independent variable is greater than for another level.

Formally, a hypothesis can be written as:
\begin{align}
    &H_0: \phi(X, C, Y) = 0\\
    &H_a: \phi(X, C, Y) \neq 0
\end{align}

where $\phi$ is a function that describes the hypothesized relationship or effect among $X$, $C$, and $Y$. The actual form of $\phi$ depends on the specific nature of the hypothesis being tested. Hypotheses are evaluated by analyzing data collected from experiments to determine whether there is sufficient evidence to reject $H_0$ in favor of $H_a$. Here statistical tests should be used as shown in section \ref{statdata}.

The hypothesis posits that manipulating or varying the independent variables $\mathbf{X}$ will lead to a change or effect in the dependent variables $\mathbf{Y}$. This relationship is testable, meaning that experimental data can potentially prove the hypothesis true or not true. In accordance with \cite{Rapp1975} this definition includes that the content of a hypothesis must enable an empirical and isolated evaluation.

In more general terms a hypothesis must fulfill the following:\\
\textbf{Clarity and Specificity}: The hypothesis should be clear and precise, making it understandable without ambiguity. It must be specific, detailing the expected relationship between independent and dependent variables.\\
\textbf{Testability and Falsifiability}: The hypothesis must be testable, meaning that it can be supported or refuted through empirical evidence obtained via experimentation or observation. It must enable the derivation of specific, testable predictions to ensure that the relationship between variables can be empirically evaluated. There must be possible empirical outcomes that could prove the hypothesis wrong.\\

These criteria ensure that a hypothesis is not only theoretically sound but also practically applicable in guiding experimental design and analysis. In the domain of CPS, research hypotheses commonly focus on comparing a novel method against established baselines or evaluating performance against a predefined threshold for a particular metric on a specific task. These hypotheses can be categorized into qualitative and quantitative types. Qualitative hypotheses aim to establish a superior/inferior relationship, assessing whether the new method outperforms or underperforms relative to existing approaches without necessarily quantifying the difference. On the other hand, quantitative hypotheses seek to define a measurable relationship, often in terms of specific performance metrics (e.g., accuracy, efficiency, response time). The majority of hypotheses are quantitative. This is because ML experiments typically generate vast amounts of data that can be precisely measured and statistically analyzed, allowing for concrete comparisons and evaluations.

\subsection{Experiment Design} \label{expdes}
The purpose of experiments is to generate empirical evidence about the validity of hypotheses. Hence, experiments need to be designed in a way that they follow the logic of hypotheses and experimental parameters need to be controlled, so that the relation of interest can be investigated in a statistically significant manner.

In ML an experiment consists of a ML-program (code) which is a concrete implementation of a ML-algorithm, run in a specific environment (hardware and software) and given specific data as input.

Following the definition of a hypothesis, let an experiment $\epsilon$ be defined as a function that maps a set of independent variables $X$ and a set of control variables $C$ to a set of dependent variables $Y$.

\begin{equation}
    \epsilon: (X, C) \rightarrow Y
\end{equation}

Both \(X\) and \(C\) can include variables from different domains. If a variable is real-valued, it belongs to the domain \( \mathbb{R} \), or a subset thereof, representing continuous values. If a variable is discrete, it is defined over \( \mathbb{Z} \), or a specific subset thereof, representing countable values. For categorical variables, we define a set \( \mathcal{C}^i = \{Category_1, Category_2, ..., Category_i\} \), where each category represents a distinct qualitative value. The dependent variable(s) \(Y\) can also be from one of these domains, depending on the nature of the outcome being measured in the experiment.

With $n,m,o,p,q,r \in \mathbb{N}_{0}$, we define the domains of \(X\), \(C\), and \(Y\) as follows:

\begin{align}
    &X \subseteq \mathbb{R}^n \cup \mathcal{C}^m\\
    &C \subseteq \mathbb{R}^o \cup \mathcal{C}^p\\
    &Y \subseteq \mathbb{R}^q \cup \mathcal{C}^r
\end{align}

The function $\epsilon$ maps combinations of independent and control variables to an outcome in \(Y\), modeling the experiment's behavior:

\begin{equation}
    \epsilon: ((\mathbb{R}^n \cup \mathcal{C}^m) \times (\mathbb{R}^o \cup \mathcal{C}^p)) \rightarrow (\mathbb{R}^q \cup \mathcal{C}^r)
\end{equation}

To illustrate this, we take a typical ML experiment, where we want to determine the anomaly detection performance of a model. Our control variables \(C\) would comprise factors like the model architecture (e.g., convolutional neural network, recurrent neural network), preprocessing techniques (e.g., normalization, dimensionality reduction), and the fixed parameters of the model (e.g., layers, activation functions) that are not subject to change during the experiment. These control variables are kept constant to ensure that the experiment accurately assesses the effect of the independent variables on the performance. The independent variables \(X\) could include the hyperparameters of the model that we intend to optimize, such as the learning rate, batch size, or the number of epochs, but also the data which is used or random seed values. These variables are systematically varied to observe their effect on the model's anomaly detection performance. Finally, the dependent variables \(Y\) would represent the outcomes of the experiment, specifically the performance metrics of the anomaly detection model, such as precision, recall, or F1 score. These metrics quantify the model's effectiveness in distinguishing between normal and anomalous instances in the dataset.

In summary, a hypothesis for this ML experiment might propose that certain configurations of the learning rate, batch size, and number of epochs (\(X\)) significantly improve the model's anomaly detection performance (\(Y\)), under fixed conditions of model architecture, preprocessing techniques, and model parameters (\(C\)). This hypothesis can then be tested through a series of experiments, analyzing the collected data to support or refute the proposed relationship.

A commonly used methodology used in this context is the Design of Experiments (DoE) method \cite{Fisher1935}, which offers a systematic approach for the selection of independent and control variables.

\subsection{Experiment Execution}
The process of conducting experiments involves translating a ML algorithm into a ML program. This program is designed to produce dependent variables from the predetermined independent and control variables. Given that a ML program operates within a complex ecosystem influenced by software and hardware dependencies, it's imperative to structure experiments to ascertain the impact of these dependencies. Additionally, it's crucial to determine the presence of any stochastic behavior in the outcomes, ensuring that the observed results are robust and reliable under varying conditions. Aleatoric and epistemic uncertainty in empirical research \cite{Huellermeier2021} must be reduced as much as possible.
To mitigate such effects, methods as seen in the state of the art can be applied. Replication runs should be used to ascertain replicable results, whilst multiple random seeds, shuffled data and hyperparameters should be tested to rule out stochastic effects. The use of multiple (standard) data sets enables a broader validity of the research. The desicion for these aspects are made during the previous design phase, but can be adapted in a exploratory manner.

We propose to design the experiment as a modular piece of software, with inputs being only the independent and control variables and outputs being the dependent variables to increase transparency and replicability of the results. The data of every experiment shall be saved in a complete and structured manner to enable the following analysis on the relation of independent and dependent variables (see \ref{documentation}, \ref{checklist}).

\subsection{Statistical Data Analysis} \label{statdata}
\begin{table*}[ht]
\centering
\caption{Comparison of statistical tests based on distribution and variance}
\label{tab:stat_tests}
\begin{tabular}{lll}
\hline
\textbf{Distribution} & \textbf{Variances} & \textbf{Appropriate Test} \\ \hline
Normal                & Equal              & Paired t-test             \\
Normal                & Unequal            & Paired t-test with Welch's correction \\
Not Normal            & Any                & Wilcoxon signed-rank test \\
Mixed                 & Any                & Mann-Whitney U test, Kolmogorov-Smirnov test \\ \hline
\end{tabular}
\end{table*}
Statistical testing \cite{Ille2008, Schiefer2021} is crucial for evaluating research hypotheses, accommodating scenarios both with and without a predefined null hypothesis (\(H_0\)). When \(H_0\) is specified, it generally asserts the absence of an effect or relationship, serving as a reference for testing the experimental data. Various statistical tests, such as t-tests, ANOVA, or chi-square tests, are utilized to compute the probability (p-value) of observing the experimental outcomes if \(H_0\) were accurate. A result is considered statistically significant if the p-value is less than a predetermined threshold (e.g., \(\alpha = 0.05\)), indicating evidence to reject \(H_0\) in favor of the alternative hypothesis (\(H_a\)), which asserts a specific effect or relationship. The choice of the appropriate test depends on the attributes of the data distributions. See table 1 for our suggestions.

In research that does not explicitly use a null hypothesis, the emphasis is on directly evaluating the support for the specific hypothesis under investigation. The statistical analysis aims to quantify the strength and reliability of the observed effects or relationships. Techniques such as confidence intervals, effect sizes, and other statistical measures are applied to assess the hypothesis' credibility. The lack of a null hypothesis underscores the importance of statistical precision in estimating and interpreting the significance and relevance of the study's findings.

Regardless of the presence of a null hypothesis, the core objective of statistical analysis in hypothesis testing is to meticulously quantify the uncertainty surrounding empirical observations. This rigorous approach ensures that conclusions derived from the data are not only based on detected patterns but also statistically substantiated. We provide additional material and code for the design and testing of hypotheses on github \cite{Vranjes2024}.

\subsection{Documentation and Publication} \label{documentation}

During the execution of the experiments data are generated. To enable unambiguous interpretation of the experimental data, all aspects of an experiment need to be documented. We propose that for every run of an experiment all independent, control and dependent variables are documented and that learned models are also saved where possible. This approach guarantees a full traceability of results. Traceable results increase replicability and offer a certain level of quality assurance.

We encourage the application of the FAIR principles \cite{Wilkinson2016} for published data to increase the usability of it. In general the documentation should also include the specifications of used software including the software version number and hardware (CPU, GPU, memory). Total needed computation time on the used environment should also be sated.

\section{Conclusion and Outlook} \label{outlook}
In this paper, we delve into the critical aspects of conducting empirical research in the ML domain, focusing on the importance of developing clear hypotheses, designing thorough experiments, executing these with precision, and conducting detailed statistical analysis. Our analysis highlights the absence of standardized methods, leading to significant variations in research quality. The process from formulating hypotheses to analyzing outcomes is fundamental for advancing ML, yet the diversity in research practices signals a need for a more unified approach.

To bridge this gap, we introduce a research process model and a set of guidelines aimed at improving the consistency, reliability, and overall impact of empirical research in ML. By adopting these recommendations, the research community can move towards more standardized practices, enhancing the replicability of studies and the reliability of their results.

This paper represents an initial effort to establish unified guidelines for high-quality empirical research within the ML field. We recognize the opportunity to further elaborate on the descriptions of process phases and artifacts, with the goal of crafting a standard process model. Such a model could serve as a foundation for ensuring research quality and facilitating its adaptation to industrial applications.

\section{Checklist} \label{checklist}

This document concludes with a comprehensive checklist that encompasses critical aspects of research methodology, including hypothesis formulation, experimental design, execution, data analysis, and documentation. The checklist serves as a systematic guide to ensure thoroughness, clarity, and adherence to scientific standards throughout the research process. It aims to facilitate the validation of the study's methodology and enhance the reliability and impact of its findings. By incorporating these considerations, researchers can bolster the integrity of their work, promoting transparency and reproducibility in the scientific community.

\begin{samepage}
\begin{itemize}[label=$\square$]
  \item Falsifiable hypotheses defined
  \item Independent variables defined
  \item Control variables defined
  \item Dependent variables defined
  \item Baseline models defined
  \item Multiple data sets selected
  \item Replication runs (re-runs) performed
  \item All random seeds set and multiple values tested
  \item Cross validations over partial data sets performed
  \item Hyperparameter tuning for every model including baselines performed
  \item Results averaged with mean and variance values over cross validations, seeds and replications
  \item Statistical testing of hypotheses performed
  \item Code published
  \item Software environment published
  \item Data published (FAIR)
  \item Trained model (weights) published
\end{itemize}
\end{samepage}

\clearpage
\printbibliography
\end{document}